\documentclass[review,5p,twocolumn]{elsarticle}

\journal{}

\usepackage{hyperref}
\usepackage{times}  
\usepackage{url}  
\usepackage{graphicx}  
\usepackage{amsmath}
\usepackage{amsfonts}
\usepackage{tabularx}
\usepackage{multirow}
\usepackage{makecell}
\usepackage{latexsym}

\usepackage[labelfont=bf,justification=raggedright,singlelinecheck=false]{caption}
\usepackage[flushleft]{threeparttable}

\usepackage[figuresright]{rotating}
\usepackage{booktabs}

\biboptions{sort&compress}

\allowdisplaybreaks[4] 

\begin{document}

\begin{frontmatter}

\title{Convolutional Gated Recurrent Units for Medical Relation Classification}
\author[a]{Bin He}
\address[a]{Research Center of Language Technology, Harbin Institute of Technology, Harbin, China}
\ead{hebin\_hit@hotmail.com}
\author[a]{Yi Guan\corref{cor}}
\cortext[cor]{Corresponding author}
\ead{guanyi@hit.edu.cn}
\author[b]{Rui Dai}
\address[b]{Department of Mathematics, Harbin Institute of Technology, Harbin, China}
\ead{13B912003@hit.edu.cn}

\begin{abstract}
Convolutional neural network (CNN) and recurrent neural network (RNN) models have become the mainstream methods for relation classification. We propose a unified architecture, which exploits the advantages of CNN and RNN simultaneously, to identify medical relations in clinical records, with only word embedding features. Our model learns phrase-level features through a CNN layer, and these feature representations are directly fed into a bidirectional gated recurrent unit (GRU) layer to capture long-term feature dependencies. We evaluate our model on two clinical datasets, and experiments demonstrate that our model performs significantly better than previous single-model methods on both datasets.
\end{abstract}


\end{frontmatter}


\section{Introduction}

\begin{figure}[!hbt]
	\centering
	\includegraphics[width=\linewidth]{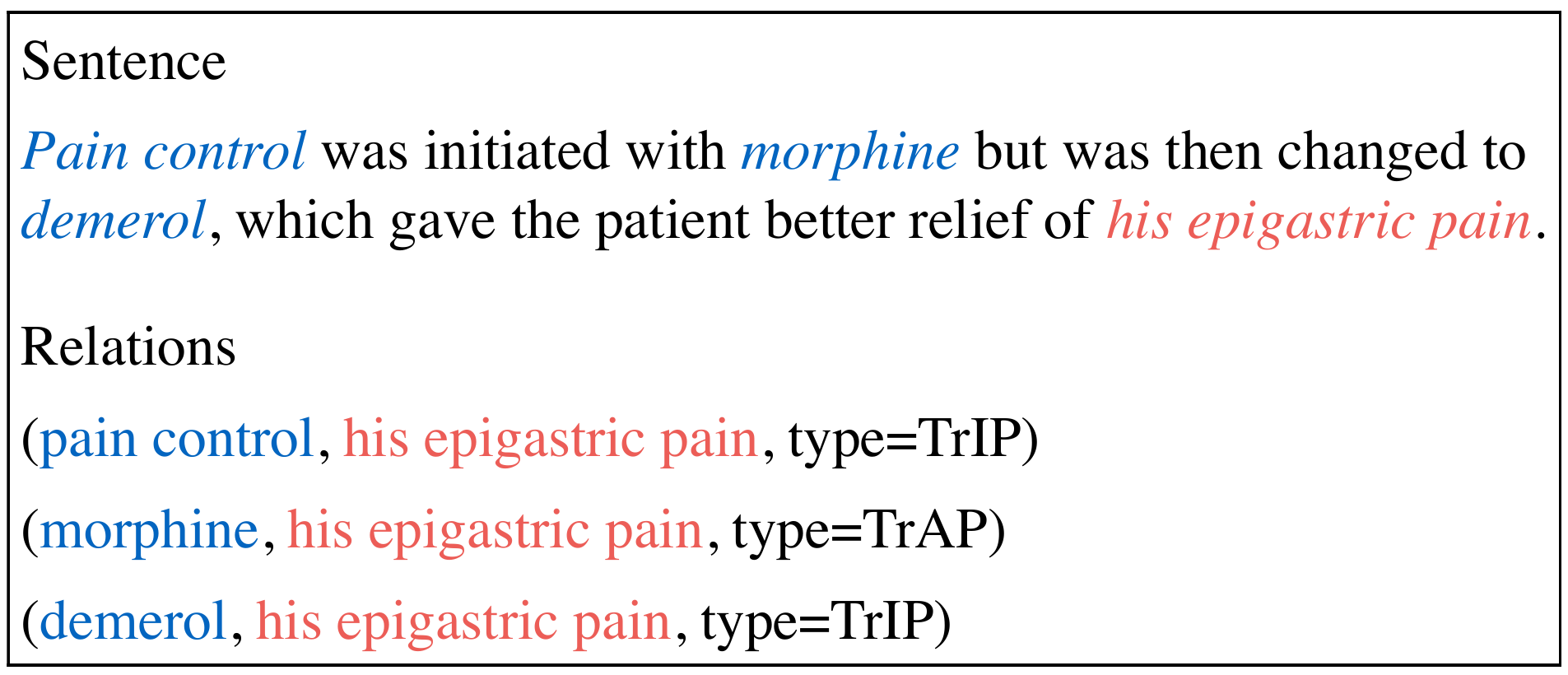}
	\caption{\label{fig:sample} An example of medical relations in a sample sentence. TrIP, treatment improves medical problem; TrAP, treatment is administered for medical problem.}
\end{figure}

Relation classification, a natural language processing (NLP) task which identifies the relation between two entities in a sentence, is an important technique in many subsequent NLP applications, such as question answering and knowledge base completion. In the clinical domain, Informatics for Integrating Biology and the Bedside (i2b2) released an annotated relation dataset on clinical records and attracted considerable attention \cite{Uzuner2011}. Identifying relations in clinical records is a challenging task because one sentence from clinical records may contain more than two medical concepts and a concept may contain several words. Figure~\ref{fig:sample} illustrates relation samples in this task.

Due to the powerful feature learning ability, convolutional neural network (CNN) and recurrent neural network (RNN) are the mainstream architectures in the relation classification task \cite{Zhang2015a,liu-EtAl:2015:ACL-IJCNLP,xu-EtAl:2015:EMNLP1,Xu2015,xu-EtAl:2016:COLING1,miwa-bansal:2016:P16-1,Wang2016,Zhou2016}. In order to utilize the advantages of these two neural networks simultaneously, combinations of CNN and RNN turn into a research trend. The most direct way is to use the voting scheme \cite{vu-EtAl:2016:N16-1}. The second combination way is to feed features extracted by a RNN architecture into CNN \cite{Cai2016,Raj2017}, which can be seen as generating new input representations by RNN. The third way is to stack RNN on CNN. Even though this architecture has not been applied to identify medical relations from clinical text, its variants have achieved remarkable results in many other classification tasks \cite{Tang2015,Nguyen2015a,zhou2015c,choi2017convolutional}.

Deep learning methods have presented satisfactory results \cite{Socher2012,yu2014factor,Zhang2015a,liu-EtAl:2015:ACL-IJCNLP,xu-EtAl:2015:EMNLP1,Xu2015,xu-EtAl:2016:COLING1,miwa-bansal:2016:P16-1} and make the models less dependent on manual feature engineering. Moreover, some researchers proposed models only with word representations as input features \cite{DBLP:conf/acl/SantosXZ15,Zhou2016}, which achieved outstanding results. Similarly, our goal is to propose a model for relation classification on clinical records, without using any external feature set. In this work, we follow the third combination way and design a two-layer architecture: input representations (word-level) are fed into a CNN layer to learn n-gram features (phrase-level), and these feature representations are directly used as the input of a bidirectional gated recurrent unit (GRU) \cite{bahdanau2014neural} layer to achieve the final sample representation (sentence-level). Our main contributions are as follows: (1) we propose a unified architecture to identify medical relations in clinical records, which has the ability to capture both local features (extracted by a CNN layer) and sequential correlations among these features (extracted by a bidirectional GRU layer); (2) we also explore training our model with attention mechanism (C-BGRU-Att) and compare the performance with the model using the conventional max-pooling operation (C-BGRU-Max); (3) experiments show our model achieves better performance than previous single-model methods, with only word embedding features.

\section{Methodology}

\begin{figure}[!hbt]
	\centering
	\includegraphics[width=\linewidth]{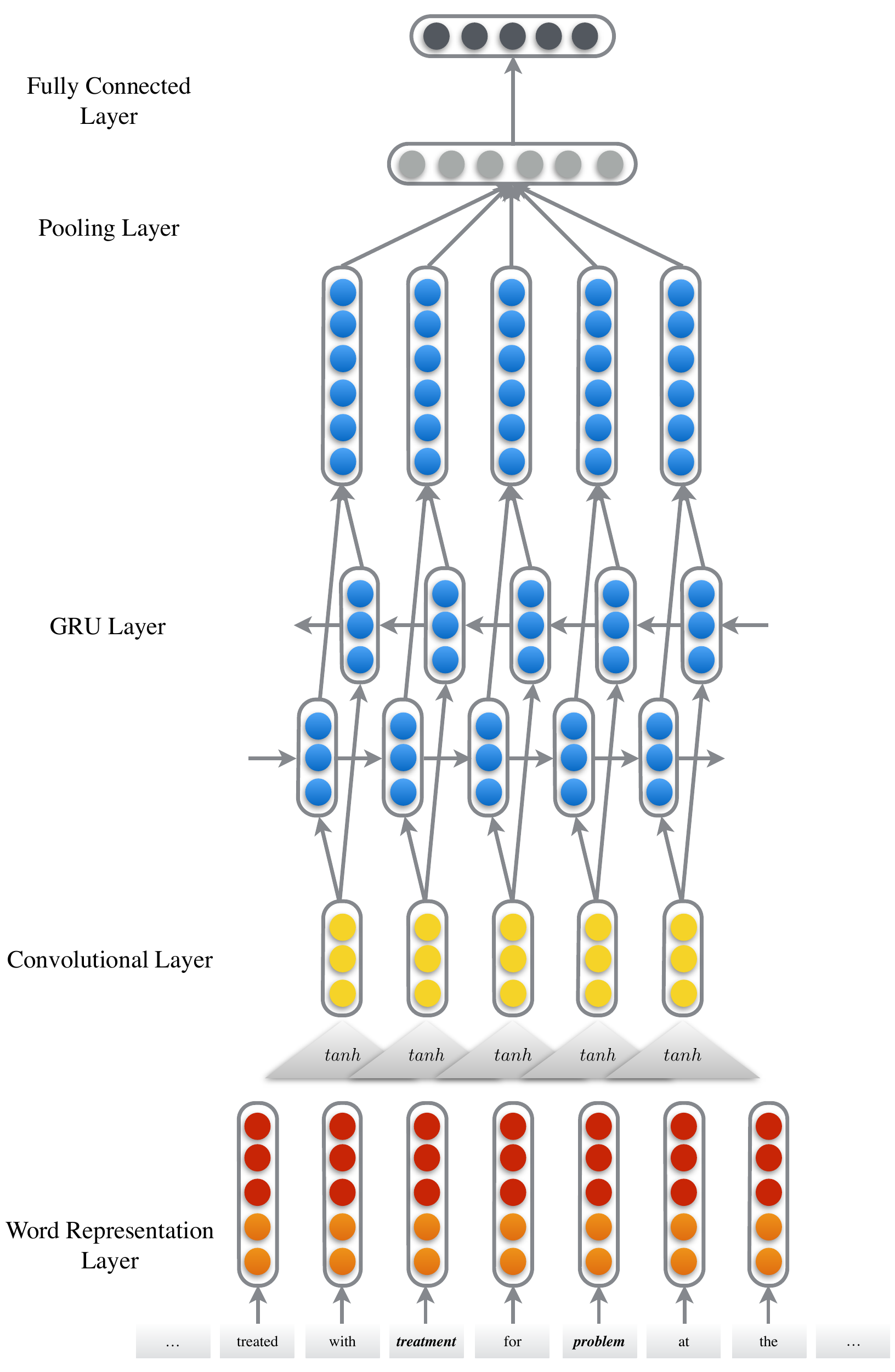}
	\caption{\label{fig:arch} Architecture of our model for medical relation classification. In the input of this architecture, concept contents in the relation sample ``she was treated with $[\text{steroids}]_{treatment}$ for $[\text{this\ swelling}]_{problem}$ at the outside hospital , and these were continued ." are replaced by their concept types.}
\end{figure}

Figure~\ref{fig:arch} describes the architecture of our model for medical relation classification on clinical records. This model learns a distributed representation for each relation sample, and calculates final scores with relation type representations. More details will be discussed in the following sections.

\subsection{Word representation layer}
\label{sec:wordrepresentation}


With reference to a previous study on relation classification \cite{Zeng2014}, word position features capture information of the relative position between words and target concepts. Therefore, an word embedding matrix $W^{w}\in\mathbb{R}^{d^w\times |V^w|}$ and an word position embedding matrix $W^{wp}\in\mathbb{R}^{d^p\times |V^p|}$ are given in this work, where $V^w$ is the vocabulary, $V^p$ is the word position set, and $d^w$ and $d^p$ are pre-set embedding sizes. Every word in the relation sample is mapped to a column vector $\mathbf{x}^w_i$ to represent the word feature. In addition, relative distances between the current word and the target concepts are mapped to word position vectors $\mathbf{x}^{p_1}_i$ and $\mathbf{x}^{p_2}_i$. Based on the above features, each word can be represented by $\mathbf{x}_i'={[{(\mathbf{x}^w_i)}^\mathrm{T}, {(\mathbf{x}^{p_1}_i)}^\mathrm{T}, {(\mathbf{x}^{p_2}_i)}^\mathrm{T}]}^\mathrm{T}$, and $\mathbf{x}_i'\in\mathbb{R}^{d^x}$, where $d^x=d^w+2d^p$.

\subsection{Convolutional layer}
\label{sec:conv}

The semantic representations of n-grams are valuable features to the relation classification task, and convolution operation can capture this information by combining word embedding features in a fixed window. 
Given the input representation $\mathbf{x}'=(\mathbf{x}_1',\mathbf{x}_2',\dots,\mathbf{x}_n')$ and a context window size $k$, concatenation of successive words in this window size can be defined as $X_j={[{\mathbf{x}_j'}^\mathrm{T},\dots,{\mathbf{x}_{j+k-1}'}^\mathrm{T}]}^\mathrm{T}$, and the representation of this relation sample can be reformatted as $X=(X_1,\dots,X_{n-k+1})$. Given a weight matrix of the convolutional filters $W^{conv}$ and a linear bias $\mathbf{b}$, the local feature representations are computed:
\begin{equation}
C_j=tanh(W^{conv}\cdot X_j + \mathbf{b}),
\end{equation}
where $W^{conv}\in\mathbb{R}^{d^c\times d^xk}$, $\mathbf{b}\in\mathbb{R}^{d^c}$, and $tanh$ denotes the hyperbolic tangent function.

Generally, this convolutional result will be fed into a max-pooling operation to extract the most significant features. However, these extracted features are independent, and the correlation information among the local features are not captured. GRU has the ability to make up for this deficiency by using a gating mechanism to capture short-term and long-term dependencies. Therefore, in this study, a GRU layer is stacked on top of the convolutional layer to continue the feature extraction work.

\subsection{GRU layer}

Similar to the long short-term memory (LSTM) unit with a memory cell and three gating units \cite{Hochreiter1997,graves2012supervised}, GRU is much simpler to compute because only two gating units are used to adaptively capture dependencies over different time scales: one is the reset gate $\mathbf{r}_j$, which controls how much information from the previous hidden state is kept in the candidate hidden state; another is the update gate $\mathbf{z}_j$, which decides how much previous information contributes and how much information from the candidate hidden state is added. The computational process are demonstrated by the following equations:
\begin{align}
\mathbf{r}_j &= \sigma(W_r\cdot C_j + U_r\cdot \mathbf{h}_{j-1} + \mathbf{b}_r), \\
\mathbf{z}_j &= \sigma(W_z\cdot C_j + U_z\cdot \mathbf{h}_{j-1} + \mathbf{b}_z), \\
\tilde{\mathbf{h}}_j &= tanh(W_h\cdot C_j + \mathbf{r}_j\odot (U_h\cdot \mathbf{h}_{j-1}) + \mathbf{b}_h), \\
\mathbf{h}_j &= (1-\mathbf{z}_j)\odot \mathbf{h}_{j-1} + \mathbf{z}_j\odot \tilde{\mathbf{h}}_j,
\end{align}
where $\sigma$ is the logistic sigmoid function, $\odot$ stands for the element-wise multiplication, $C_j$ is the current n-gram feature representation (mentioned in Section~\ref{sec:conv}), $\mathbf{h}_{j-1}$ and $\tilde{\mathbf{h}}_j$ are the previous and the candidate hidden state, respectively, and $\mathbf{h}_j\in\mathbb{R}^{d^h}$ is the current hidden state. 
$W_r$, $U_r$, $\mathbf{b}_r$, $W_z$, $U_z$, $\mathbf{b}_z$, $W_h$, $U_h$ and $\mathbf{b}_h$ are weight matrices to be learned.

We use a bidirectional GRU \cite{bahdanau2014neural} to encode the n-gram feature representations, which contains a forward GRU and a backword GRU. 
A sequence of forward hidden states $(\overrightarrow{\mathbf{h}}_1, \dots, \overrightarrow{\mathbf{h}}_{n-k+1})$ and a sequence of backward hidden states $(\overleftarrow{\mathbf{h}}_1, \dots, \overleftarrow{\mathbf{h}}_{n-k+1})$ are obtained. The final $j$-th hidden state can be achieved by concatenating the $j$-th forward and backward hidden state: $\mathbf{h}_j = {[{\overrightarrow{\mathbf{h}}_j}^\mathrm{T}, {\overleftarrow{\mathbf{h}}_j}^\mathrm{T}]}^\mathrm{T}$, which contains the dependencies of the preceding and the following n-gram features.

\subsection{Pooling layer}

Two different kinds of pooling schemes are adopted to generate the semantic representation of the relation sample $\mathbf{rs}$. 

{\bf Max pooling} can be seen as a down-sampling operation that aims to extract the most significant features. After using this operation in our network, the $i$-th feature value $\mathbf{rs}_i$ is calculated by
\begin{equation}
\mathbf{rs}_i = max([\mathbf{h}_1]_i, \dots, [\mathbf{h}_{n-k+1}]_i),
\end{equation}
where $[\mathbf{h}_j]_i$ denotes the $i$-th element in vector $\mathbf{h}_j$. And all these features constitute the semantic representation of the relation sample $\mathbf{rs} = {(\mathbf{rs}_1, \dots, \mathbf{rs}_{d^h})}^\mathrm{T}$.

{\bf Attentive pooling} Given the output of the GRU layer $H=[\mathbf{h}_{1}, \dots, \mathbf{h}_{n-k+1}]$, we follow the attention mechanism used in \cite{Zhou2016}, and the representation $\mathbf{rs}$ is formed:
\begin{align}
\boldsymbol{\alpha} &= softmax(\mathbf{v}^\mathrm{T}\cdot tanh(H)), \\
\mathbf{rs} &= tanh(H\cdot {\boldsymbol{\alpha}}^\mathrm{T}),
\end{align}
where $\mathbf{v}$ is a model parameter vector and $\boldsymbol{\alpha}$ is a weight vector to measure which parts of the GRU output are relatively significant for the relation classification.

\subsection{Fully connected layer}

We apply a softmax classifier to achieve the confidence scores with a class embedding matrix $W^{cs}$:
\begin{equation}
\mathbf{s}_{\theta} = softmax(W^{cs}\cdot \mathbf{rs}),
\end{equation}
where $\theta$ is the model parameter set. $\mathbf{s}_{\theta}^y$ is the confidence score of the true relation type $y$, and the loss function can be defined as
\begin{equation}
\mathcal{L} = -\frac{1}{m}\sum_{i=1}^{m}{\log \mathbf{s}_{\theta}^y} + \beta||\theta||^2,
\end{equation}
where $m$ is the sample size and $\beta$ is the $l_2$ regularization parameter.

\section{Experiments}
\subsection{Dataset and experimental settings}
Experiments are conducted on the 2010 i2b2/VA relation dataset\footnote{The relation dataset is available at \url{https://www.i2b2.org/NLP/Relations/}.} and the WI relation dataset\footnote{\url{https://github.com/WILAB-HIT/Resources/tree/master/entity_assertion_relation}}. The former dataset comprises 426 English discharge summaries (170 for training and 256 for test), and the latter dataset contains 992 Chinese clinical records (521 for training and 471 for test). The relation types and their counts in these two datasets are listed in Table~\ref{tab:relationcount}. As stipulated in the official evaluation metric in the 2010 i2b2/VA challenge, the model performance is based on the micro-averaged F1 score over all positive relation types.

\begin{table}[!htb]
	\footnotesize
	\caption{\label{tab:relationcount}Relation type statistics.}
	\begin{threeparttable}
		\begin{tabularx}{\linewidth}{XXXXXX}
			\hline
			\multicolumn{3}{c}{2010 i2b2/VA relation dataset} & \multicolumn{3}{c}{WI Relation dataset} \\ \cmidrule(lr){1-3} \cmidrule(lr){4-6}
			Relation & Train &  Test &  Relation & Train &  Test \\ \hline
			TrIP &  51 & 152 & TrID & 103 & 92\\
			TrWP & 24  &  109  & TrWD & 38 & 27 \\ 
			TrCP & 184 & 342 &  TrAD & 221 & 166 \\
			TrAP  & 885 & 1732  &  NTrD & 675 & 656 \\
			TrNAP & 62 & 112 & TrIS & 337 & 215 \\
			NTrP & 1702 & 2759 &  TrWS & 297 & 242 \\ 
			TeRP & 993 & 2060 &  TrCS & 125 & 176 \\
			TeCP & 166 & 338 & TrAS & 334 & 238 \\
			NTeP & 993  & 1974 &  NTrS & 1062 & 901 \\
			PIP & 755 & 1448  & TeRD & 301 & 227 \\
			NPP & 4418 & 8089  & NTeD & 331 & 248 \\ \cmidrule(lr){1-3} \morecmidrules\cmidrule(lr){1-3}
			SID & 969 & 620 & TeRS & 527 & 542 \\
			DCS & 228 & 181 & TeAS & 313 & 564 \\
			NDS & 777 & 635 & NTeS & 8628 & 7060 \\ \hline
		\end{tabularx}
		\begin{tablenotes}[para,flushleft]
			Positive relations were annotated in both relation datasets, and samples of negative relation types (starting with ``N" in this table) were extracted to ensure each concept pair within a sentence could be assigned a certain relation type. For more details of these relation types, please refer to \cite{Uzuner2011,He2016}.
		\end{tablenotes}
	\end{threeparttable}
\end{table}


In our methods, the initial word representations and the other matrices are randomly initialized by normalized initialization \cite{DBLP:journals/jmlr/GlorotB10}, and a 5-fold cross-validation is used on the training set to tune the model hyperparameters. The selected hyperparameter values are: word embedding size $d^w$, 100; word position embedding size $d^p$, 10; convolutional size $d^c$, 200; context window size $k$, 3; GRU dimension $d^h$, 100; learning rate, 0.01. 
Adam technique \cite{kingma2014adam} is utilized to optimize our loss function. We use both $l_2$ regularization and dropout technique \cite{Srivastava2014} to avoid overfitting, and the values are set to 0.0001 and 0.5, respectively.

\subsection{Baselines}
\subsubsection{2010 i2b2/VA relation dataset} 
When doing experiments on this dataset, the previous methods \cite{DBLP:conf/coling/DSouzaN14,Sahu2016Relation,Raj2017} followed inconsistent data split schemes. In order to compare these methods together, we choose the split scheme in \cite{DBLP:conf/coling/DSouzaN14}, which is also the official data split.

{\bf SVM}: due to the dataset available to the research community is only a subset of the dataset used in the 2010 i2b2/VA challenge, so \citet{DBLP:conf/coling/DSouzaN14} reimplemented the state-of-the-art model in the challenge \cite{Rink2011} and reevaluated this model on the relation dataset accessible. {\bf SVM+ILP}: \citet{DBLP:conf/coling/DSouzaN14} also proposed a better single-model method and an ensemble-based method within an integer linear programming (ILP) framework. In these feature-based state-of-the-art methods, a variety of external features sets are used, such as part-of-speech (POS) tagging and dependency parsing.

In this work, three previous neural network methods are reimplemented and reevaluated. {\bf CNN}: a multiple-filter CNN with max-pooling proposed by \citet{Sahu2016Relation}. To evaluate the model performance independent of the external features, POS and chunk features used in this method are removed. {\bf CRNN-Max and CRNN-Att}: a two-layer model comprising recurrent and convolutional layers with max and attentive pooling \cite{Raj2017}. However, only word embeddings were used in their work. In order to maintain a fair comparison, word position embeddings are added in our model reimplementation. In these three baseline reimplementations, we follow the selected hyperparameters used in the corresponding work and the word embeddings are pre-trained on the deidentified notes from the MIMIC-III database \cite{Johnson2016}.

\subsubsection{WI relation dataset} 
{\bf SVM}: this model is implemented using scikit-learn\footnote{\url{http://scikit-learn.org/stable/}.}. And it involves the following features: entity $e_1$, entity $e_2$, entity type $et_1$, entity type $et_2$, distance between $e_1$ and $e_2$, words in $e_1$ and $e_2$, words between $e_1$ and $e_2$, words behind $e_2$, POS of words in $e_1$ and $e_2$, POS of words between $e_1$ and $e_2$, and POS of words behind $e_2$.

{\bf CNN}: the model version of C-BGRU-Max after removing the GRU layer, which is a CNN-based model.

\subsection{Experimental results}
\subsubsection{System performance}
The performance results are displayed in Table~\ref{tab:classifiers} and \ref{tab:classifiers_ct}, including 95\% confidence intervals for the models we implemented, which are derived using bootstrapping \cite{DiCiccio1996}. We use the same bootstrapping method described in \cite{gao2017hierarchical}. We observe that our C-BGRU-Max model outperforms the previous single-model methods significantly in both datasets, without using any external features. After using attentive pooling, the model performance on the two datasets shows different changes: drops on the 2010 i2b2/VA relation dataset but increases on the WI relation dataset. The intuitive explanation is that descriptions in English discharge summaries tend to be more colloquial, making specific features more difficult to capture. More details of the category-wise and class-wise performance comparisons are listed in Table~\ref{tab:category}, \ref{tab:subclasses}, \ref{tab:category_ct}, and \ref{tab:subclasses_ct}.

\begin{table}[!htb]
	\centering
	\footnotesize
	\caption{\label{tab:classifiers}System performance comparison with other models using the 2010 i2b2/VA relation dataset.}
	\begin{threeparttable}
		\begin{tabularx}{\linewidth}{llXXX}
			\hline Classifier & \makecell[l]{External\\features} & P & R & F1 \\ \hline
			\multicolumn{5}{l}{\emph{Single-model methods}} \\
			SVM$^*$ \cite{Rink2011} & Set1 & 58.1 & \bf 66.7 & 62.1 \\
			SVM+ILP \cite{DBLP:conf/coling/DSouzaN14} & Set2 & \bf 75.0 & 58.9 & 66.0 \\
			\makecell[l]{CNN \cite{Sahu2016Relation}\\$ $} & \makecell[l]{None\\$ $} & \makecell[l]{68.0\\(67.4, 68.6)} & \makecell[l]{55.1\\(54.5, 55.7)} & \makecell[l]{60.9\\(60.4, 61.4)} \\
			\makecell[l]{CRNN-Max \cite{Raj2017}\\$ $} & \makecell[l]{None\\$ $} & \makecell[l]{65.1\\(64.6, 65.6)} & \makecell[l]{61.3\\(60.7, 61.8)} & \makecell[l]{63.1\\(62.7, 63.6)} \\
			\makecell[l]{CRNN-Att \cite{Raj2017}\\$ $} & \makecell[l]{None\\$ $} & \makecell[l]{63.2\\(62.6, 63.7)} & \makecell[l]{58.5\\(58.0, 59.0)} & \makecell[l]{60.7\\(60.3, 61.2)} \\
			\makecell[l]{C-BGRU-Max\\$ $} & \makecell[l]{None\\$ $} & \makecell[l]{69.3\\(68.8, 69.9)} & \makecell[l]{66.3\\(65.8, 66.8)} & \makecell[l]{\bf 67.8\\(67.3, 68.3)} \\
			\makecell[l]{C-BGRU-Att\\$ $} & \makecell[l]{None\\$ $} & \makecell[l]{69.6\\(69.0, 70.1)} & \makecell[l]{63.7\\(63.1, 64.2)} & \makecell[l]{66.5\\(66.0, 66.9)} \\  \hline 
			\multicolumn{5}{l}{\emph{Ensemble-based method}} \\
			Ensemble+ILP$^{\circ}$ \cite{DBLP:conf/coling/DSouzaN14} & Set2 & 66.7 & 72.9 & 69.6 \\ \hline
		\end{tabularx}
		\begin{tablenotes}[para,flushleft]
			The symbol $*$ indicates that this model is reimplemented by \cite{DBLP:conf/coling/DSouzaN14} on the relation dataset available to the research community, due to the accessible dataset is only a subset of that used in the 2010 i2b2/VA challenge. The symbol $\circ$ indicates that this classifier is the ensemble of 5 independent models. The bold item is the best result. Set1: POS, chunk, semantic role labeler, word lemma, dependency parse, assertion type, sentiment category, Wikipedia. Set2: POS, chunk, semantic role labeler, word lemma, dependency parse, assertion type, sentiment category, Wikipedia, manually labeled patterns. POS, part-of-speech; ILP, integer linear programming.
		\end{tablenotes}
	\end{threeparttable}
\end{table}

\begin{table}[!htb]
	\centering
	\footnotesize
	\caption{\label{tab:classifiers_ct}System performance comparison using the WI relation dataset.}
	\begin{threeparttable}
		\begin{tabularx}{\linewidth}{lXXX}
			\hline Classifier & P & R & F1 \\ \hline
			SVM & 72.9 & 63.9 & 68.1 \\
			\makecell[l]{CNN\\$ $} & \makecell[l]{72.7\\(72.0, 73.4)} & \makecell[l]{64.5\\(63.7, 65.2)} & \makecell[l]{68.3\\(67.7, 69.0)} \\
			\makecell[l]{C-BGRU-Max\\$ $} & \makecell[l]{73.2\\(72.5, 73.9)} & \makecell[l]{68.3\\(67.6, 69.0)} & \makecell[l]{70.7\\(70.1, 71.3)} \\
			\makecell[l]{C-BGRU-Att\\$ $} & \makecell[l]{\bf 74.8\\(74.1, 75.5)} & \makecell[l]{\bf 68.8\\(68.1, 69.5)} & \makecell[l]{\bf 71.6\\(71.0, 72.3)} \\  \hline 
		\end{tabularx}
		\begin{tablenotes}[para,flushleft]
			The bold item is the best result.
		\end{tablenotes}
	\end{threeparttable}
\end{table}

\begin{table}[!htb]
	\centering
	\footnotesize
	\caption{\label{tab:category}Category-wise performance comparison with other neural network models using the 2010 i2b2/VA relation dataset.}
	\begin{threeparttable}
		\begin{tabularx}{\linewidth}{lXXXXXXXXX}
			\hline Classifier & \multicolumn{3}{l}{TrP relations} & \multicolumn{3}{l}{TeP relations} & \multicolumn{3}{l}{PP relations} \\ \cmidrule(lr){2-4}\cmidrule(lr){5-7}\cmidrule(lr){8-10}
			& P & R & F1 & P & R & F1 & P & R & F1 \\ 
			& CI($\pm$) & CI($\pm$) & CI($\pm$) & CI($\pm$) & CI($\pm$) & CI($\pm$) & CI($\pm$) & CI($\pm$) & CI($\pm$) \\ \hline
			CNN \cite{Sahu2016Relation} & 60.9 & 48.2 & 53.8 & 75.8 & 69.2 & 72.3 & 64.8 & 43.3 & 51.9 \\
			& 1.0 & 0.9 & 0.9 & 0.9 & 0.8 & 0.7 & 1.3 & 1.1 & 1.1 \\
			CRNN-Max \cite{Raj2017} & 58.4 & 53.8 & 56.0 & 73.3 & 73.1 & 73.2 & 61.6 & 54.4 & 57.8 \\
			& 0.9 & 0.9 & 0.8 & 0.8 & 0.8 & 0.7 & 1.1 & 1.2 & 1.0 \\
			CRNN-Att \cite{Raj2017} & 55.2 & 50.8 & 52.9 & 70.1 & 73.8 & 71.9 & 63.3 & 46.3 & 53.5 \\
			& 0.9 & 0.9 & 0.8 & 0.8 & 0.8 & 0.7 & 1.3 & 1.2 & 1.1 \\
			C-BGRU-Max & 62.7 & 59.7 & \underline{\bf 61.2} & 78.4 & 77.5 & \underline{\bf 77.9} & 64.8 & 58.9 & \underline{\bf 61.7} \\
			& 0.9 & 0.9 & 0.8 & 0.8 & 0.8 & 0.6 & 1.2 & 1.1 & 1.0 \\
			C-BGRU-Att & 63.9 & 57.1 & \underline{60.4} & 79.4 & 72.5 & \underline{75.8} & 62.8 & 60.0 & \underline{61.4} \\
			& 0.9 & 0.9 & 0.8 & 0.7 & 0.8 & 0.7 & 1.1 & 1.2 & 1.0 \\
			\hline
		\end{tabularx}
		\begin{tablenotes}[para,flushleft]
			TrP, Treatment-Problem; TeP, Test-Problem; PP, Problem-Problem. CI($\pm$) is confidence interval for P, R, and F1. The bold item is the best result. Compared with previous models, the underlined item is statistically significant.
		\end{tablenotes}
	\end{threeparttable}
\end{table}

\begin{sidewaystable}
	\centering
	\footnotesize
	\caption{\label{tab:subclasses}Class-wise performance comparison with other neural network models using the 2010 i2b2/VA relation dataset.}
	\begin{threeparttable}
		\begin{tabular}{lp{0.5cm}p{0.5cm}p{0.5cm}p{0.5cm}p{0.5cm}p{0.5cm}p{0.5cm}p{0.5cm}p{0.5cm}p{0.5cm}p{0.5cm}p{0.5cm}p{0.5cm}p{0.5cm}p{0.5cm}p{0.5cm}p{0.5cm}p{0.5cm}p{0.5cm}p{0.5cm}p{0.5cm}p{0.5cm}p{0.5cm}p{0.5cm}}
			\hline Classifier & \multicolumn{3}{l}{TrIP} & \multicolumn{3}{l}{TrWP} & \multicolumn{3}{l}{TrCP} & \multicolumn{3}{l}{TrAP} & \multicolumn{3}{l}{TrNAP} & \multicolumn{3}{l}{TeRP} & \multicolumn{3}{l}{TeCP} & \multicolumn{3}{l}{PIP} \\ \cmidrule(lr){2-4}\cmidrule(lr){5-7}\cmidrule(lr){8-10}\cmidrule(lr){11-13}\cmidrule(lr){14-16}\cmidrule(lr){17-19}\cmidrule(lr){20-22}\cmidrule(lr){23-25}
			& P & R & F1 & P & R & F1 & P & R & F1 & P & R & F1 & P & R & F1 & P & R & F1 & P & R & F1 & P & R & F1 \\
			& CI($\pm$) & CI($\pm$) & CI($\pm$) & CI($\pm$) & CI($\pm$) & CI($\pm$) & CI($\pm$) & CI($\pm$) & CI($\pm$) & CI($\pm$) & CI($\pm$) & CI($\pm$) & CI($\pm$) & CI($\pm$) & CI($\pm$) & CI($\pm$) & CI($\pm$) & CI($\pm$) & CI($\pm$) & CI($\pm$) & CI($\pm$) & CI($\pm$) & CI($\pm$) & CI($\pm$) \\ \hline
			CNN \cite{Sahu2016Relation} & 18.7 & 5.9 & 9.0 & 0.0 & 0.0 & 0.0 & 56.6 & 30.1 & 39.3 & 63.0 & 61.1 & 62.0 & 35.9 & 9.1 & \bf 14.5 & 77.7 & 77.8 & 77.7 & 45.0 & 16.9 & 24.5 & 64.8 & 43.3 & 51.9 \\
			& 5.0 & 1.7 & 2.5 & 0.0 & 0.0 & 0.0 & 3.4 & 2.2 & 2.4 & 1.0 & 1.0 & 0.9 & 7.5 & 2.3 & 3.5 & 0.8 & 0.8 & 0.6 & 3.9 & 1.7 & 2.3 & 1.3 & 1.1 & 1.1 \\
			CRNN-Max \cite{Raj2017} & 33.0 & 4.6 & 8.1 & 0.0 & 0.0 & 0.0 & 41.9 & 28.1 & 33.6 & 61.3 & 69.8 & 65.3 & 11.7 & 3.2 & 5.0 & 76.9 & 80.1 & 78.5 & 41.8 & 30.4 & 35.2 & 61.6 & 54.4 & 57.8 \\
			& 8.9 & 1.5 & 2.6 & 0.0 & 0.0 & 0.0 & 2.9 & 2.1 & 2.3 & 1.0 & 1.0 & 0.8 & 5.1 & 1.4 & 2.2 & 0.8 & 0.8 & 0.6 & 2.8 & 2.2 & 2.2 & 1.1 & 1.2 & 1.0 \\
			CRNN-Att \cite{Raj2017} & 0.0 & 0.0 & 0.0 & 0.0 & 0.0 & 0.0 & 34.1 & 10.8 & 16.4 & 56.4 & 69.6 & 62.3 & 0.0 & 0.0 & 0.0 & 71.5 & 83.0 & 76.8 & 45.7 & 17.8 & 25.6 & 63.3 & 46.3 & 53.5 \\
			& 0.0 & 0.0 & 0.0 & 0.0 & 0.0 & 0.0 & 3.7 & 1.4 & 2.0 & 1.0 & 1.0 & 0.8 & 0.0 & 0.0 & 0.0 & 0.8 & 0.7 & 0.6 & 3.9 & 1.9 & 2.4 & 1.3 & 1.2 & 1.1 \\
			C-BGRU-Max & 51.4 & 4.7 & 8.7 & 36.4 & 0.7 & \underline{1.4} & 52.1 & 38.9 & \underline{\bf 44.5} & 64.4 & 75.8 & \underline{\bf 69.6} & 38.5 & 7.1 & 12.0 & 79.9 & 84.4 & \underline{\bf 82.1} & 61.1 & 35.3 & \underline{\bf 44.8} & 64.8 & 58.9 & \underline{\bf 61.7} \\
			& 12.0 & 1.5 & 2.7 & 29.2 & 0.7 & 1.3 & 2.7 & 2.3 & 2.1 & 1.0 & 0.9 & 0.8 & 9.7 & 2.1 & 3.4 & 0.8 & 0.7 & 0.6 & 3.1 & 2.3 & 2.4 & 1.2 & 1.1 & 1.0 \\
			C-BGRU-Att & 43.8 & 11.1 & \underline{\bf 17.6} & 26.7 & 2.9 & \underline{\bf 5.3} & 48.5 & 41.1 & \underline{\bf 44.5} & 67.7 & 70.9 & \underline{69.3} & 30.1 & 8.8 & 13.6 & 81.2 & 79.3 & \underline{80.3} & 58.7 & 30.8 & \underline{40.4} & 62.8 & 60.0 & \underline{61.4} \\
			& 7.0 & 2.2 & 3.3 & 10.9 & 1.4 & 2.4 & 2.5 & 2.2 & 2.0 & 0.9 & 1.0 & 0.8 & 7.0 & 2.3 & 3.4 & 0.8 & 0.8 & 0.6 & 3.2 & 2.2 & 2.4 & 1.1 & 1.2 & 1.0 \\
			\hline
		\end{tabular}
		\begin{tablenotes}[para,flushleft]
			CI($\pm$) is confidence interval for P, R, and F1. The bold item is the best result. Compared with previous models, the underlined item is statistically significant.
		\end{tablenotes}
	\end{threeparttable}
\end{sidewaystable}

\begin{table*}[!htb]
	\centering
	\footnotesize
	\caption{\label{tab:category_ct}Category-wise performance of neural network models using the WI relation dataset.}
	\begin{threeparttable}
		\begin{tabularx}{\linewidth}{lXXXXXXXXXXXXXXX}
			\hline Classifier & \multicolumn{3}{l}{TrD relations} & \multicolumn{3}{l}{TrS relations} & \multicolumn{3}{l}{TeD relations} & \multicolumn{3}{l}{TeS relations} & \multicolumn{3}{l}{DS relations} \\ \cmidrule(lr){2-4}\cmidrule(lr){5-7}\cmidrule(lr){8-10}\cmidrule(lr){11-13}\cmidrule(lr){14-16}
			& P & R & F1 & P & R & F1 & P & R & F1 & P & R & F1 & P & R & F1 \\ 
			& CI($\pm$) & CI($\pm$) & CI($\pm$) & CI($\pm$) & CI($\pm$) & CI($\pm$) & CI($\pm$) & CI($\pm$) & CI($\pm$) & CI($\pm$) & CI($\pm$) & CI($\pm$) & CI($\pm$) & CI($\pm$) & CI($\pm$) \\ \hline
			CNN & 59.5 & 56.3 & 57.8 & 58.1 & 50.5 & 54.1 & 90.1 & 90.7 & 90.4 & 86.9 & 59.0 & 70.3 & 72.5 & 82.7 & 77.3 \\
			& 2.6 & 2.7 & 2.5 & 1.5 & 1.5 & 1.3 & 1.7 & 1.7 & 1.2 & 1.1 & 1.3 & 1.1 & 1.3 & 1.2 & 1.0 \\
			C-BGRU-Max & 59.9 & 60.4 & 60.1 & 59.3 & 54.2 & 56.7 & 91.7 & 92.0 & 91.8 & 85.5 & 64.7 & \underline{73.7} & 73.6 & 84.7 & \bf 78.8 \\
			& 2.4 & 2.5 & 2.3 & 1.5 & 1.5 & 1.3 & 1.6 & 1.5 & 1.1 & 1.1 & 1.3 & 1.0 & 1.2 & 1.1 & 1.0 \\
			C-BGRU-Att & 61.8 & 58.7 & \bf 60.2 & 64.4 & 55.5 & \underline{\bf 59.6} & 91.6 & 93.6 & \bf 92.5 & 82.6 & 68.0 & \underline{\bf 74.6} & 75.0 & 80.8 & 77.8 \\
			& 2.6 & 2.6 & 2.5 & 1.5 & 1.5 & 1.4 & 1.6 & 1.4 & 1.1 & 1.1 & 1.2 & 0.9 & 1.3 & 1.2 & 1.0 \\
			\hline
		\end{tabularx}
		\begin{tablenotes}[para,flushleft]
			TrD, Treatment-Disease; TrS, Treatment-Symptom; TeD, Test-Disease; TeS, Test-Symptom; DS, Disease-Symptom. CI($\pm$) is confidence interval for P, R, and F1.  The bold item is the best result. Compared with CNN, the underlined item is statistically significant.
		\end{tablenotes}
	\end{threeparttable}
\end{table*}

\begin{table*}[!htb]
	\centering
	\footnotesize
	\caption{\label{tab:subclasses_ct}Class-wise performance of neural network models using the WI relation dataset.}
	\begin{threeparttable}
		\begin{tabularx}{\linewidth}{lXXXXXXXXXXXXXXXXXX}
			\hline Classifier & \multicolumn{3}{l}{TrID} & \multicolumn{3}{l}{TrWD} & \multicolumn{3}{l}{TrAD} & \multicolumn{3}{l}{TrIS} & \multicolumn{3}{l}{TrWS} & \multicolumn{3}{l}{TrCS} \\ \cmidrule(lr){2-4}\cmidrule(lr){5-7}\cmidrule(lr){8-10}\cmidrule(lr){11-13}\cmidrule(lr){14-16}\cmidrule(lr){17-19}
			& P & R & F1 & P & R & F1 & P & R & F1 & P & R & F1 & P & R & F1 & P & R & F1 \\ 
			& CI($\pm$) & CI($\pm$) & CI($\pm$) & CI($\pm$) & CI($\pm$) & CI($\pm$) & CI($\pm$) & CI($\pm$) & CI($\pm$) & CI($\pm$) & CI($\pm$) & CI($\pm$) & CI($\pm$) & CI($\pm$) & CI($\pm$) & CI($\pm$) & CI($\pm$) & CI($\pm$) \\ \hline
			CNN & 53.7 & 39.6 & 45.6 & 53.4 & 28.9 & 37.5 & 62.0 & 70.0 & 65.8 & 50.6 & 43.7 & 46.9 & 71.5 & 69.0 & 70.2 & 69.0 & 20.2 & 31.3 \\
			& 5.4 & 4.7 & 4.5 & 11.7 & 7.6 & 8.4 & 3.1 & 3.2 & 2.6 & 3.0 & 3.0 & 2.6 & 2.5 & 2.7 & 2.0 & 5.5 & 2.7 & 3.5 \\
			C-BGRU-Max & 53.7 & 39.1 & 45.3 & 56.6 & 44.4 & \bf 49.8 & 62.3 & 74.7 & \bf 67.9 & 54.3 & 42.0 & 47.3 & 75.3 & 70.5 & \bf 72.8 & 57.5 & 27.7 & 37.4 \\
			& 5.4 & 4.4 & 4.2 & 9.5 & 8.6 & 8.0 & 2.9 & 3.0 & 2.4 & 3.4 & 3.0 & 2.7 & 2.5 & 2.6 & 2.0 & 4.7 & 2.9 & 3.3 \\
			C-BGRU-Att & 57.9 & 40.7 & \bf 47.8 & 58.7 & 40.0 & 47.6 & 63.5 & 71.7 & 67.3 & 58.8 & 50.0 & \underline{\bf 54.1} & 75.8 & 69.5 & 72.5 & 66.4 & 26.2 & \bf 37.6 \\
			& 5.5 & 4.6 & 4.4 & 10.4 & 8.4 & 8.1 & 3.1 & 3.1 & 2.6 & 3.2 & 3.0 & 2.7 & 2.5 & 2.6 & 2.1 & 5.0 & 2.9 & 3.4 \\ \cmidrule{2-19}
			& \multicolumn{3}{l}{TrAS} & \multicolumn{3}{l}{TeRD} & \multicolumn{3}{l}{TeRS} & \multicolumn{3}{l}{TeAS} & \multicolumn{3}{l}{DCS} & \multicolumn{3}{l}{SID} \\ \cmidrule(lr){2-4}\cmidrule(lr){5-7}\cmidrule(lr){8-10}\cmidrule(lr){11-13}\cmidrule(lr){14-16}\cmidrule(lr){17-19}
			& P & R & F1 & P & R & F1 & P & R & F1 & P & R & F1 & P & R & F1 & P & R & F1 \\ 
			& CI($\pm$) & CI($\pm$) & CI($\pm$) & CI($\pm$) & CI($\pm$) & CI($\pm$) & CI($\pm$) & CI($\pm$) & CI($\pm$) & CI($\pm$) & CI($\pm$) & CI($\pm$) & CI($\pm$) & CI($\pm$) & CI($\pm$) & CI($\pm$) & CI($\pm$) & CI($\pm$) \\ \cmidrule{2-19}
			CNN & 50.2 & 60.3 & 54.8 & 90.1 & 90.7 & 90.4 & 88.8 & 84.1 & \bf 86.4 & 82.9 & 35.0 & 49.2 & 56.6 & 63.2 & 59.7 & 77.0 & 88.4 & 82.3 \\
			& 2.6 & 2.8 & 2.3 & 1.7 & 1.7 & 1.2 & 1.2 & 1.3 & 0.9 & 2.1 & 1.8 & 1.9 & 3.1 & 3.2 & 2.6 & 1.4 & 1.2 & 1.0 \\
			C-BGRU-Max & 51.1 & 68.3 & 58.5 & 91.7 & 92.0 & 91.8 & 85.4 & 84.7 & 85.1 & 85.6 & 45.4 & \underline{59.4} & 56.3 & 69.0 & \bf 62.0 & 79.1 & 89.4 & \bf 83.9 \\
			& 2.4 & 2.7 & 2.1 & 1.6 & 1.5 & 1.1 & 1.4 & 1.4 & 1.0 & 1.7 & 1.9 & 1.7 & 2.8 & 3.0 & 2.4 & 1.3 & 1.1 & 0.9 \\
			C-BGRU-Att & 58.6 & 67.7 & \underline{\bf 62.8} & 91.6 & 93.6 & \bf 92.5 & 80.6 & 86.0 & 83.2 & 86.2 & 50.7 & \underline{\bf 63.9} & 58.7 & 61.7 & 60.2 & 79.6 & 86.4 & 82.9 \\
			& 2.6 & 2.6 & 2.2 & 1.6 & 1.4 & 1.1 & 1.4 & 1.3 & 1.0 & 1.6 & 1.9 & 1.7 & 3.2 & 3.2 & 2.7 & 1.4 & 1.3 & 1.0 \\
			\hline
		\end{tabularx}
		\begin{tablenotes}[para,flushleft]
			CI($\pm$) is confidence interval for P, R, and F1.  The bold item is the best result. Compared with CNN, the underlined item is statistically significant.
		\end{tablenotes}
	\end{threeparttable}
\end{table*}

\subsubsection{Discussion of attentive pooling}
As show in Table~\ref{tab:classifiers}, the F1 scores of CRNN-Att and C-BGRU-Att are lower than that of CRNN-Max and C-BGRU-Max, respectively. This indicates that the attention mechanism, which presents a positive effect in the general domain \cite{Zhou2016,kim2017multiple}, does not show any performance improvement on the 2010 i2b2/VA relation dataset. In this dataset, there exist $\sim$3.3 entities in each sentence on average. Therefore, input representations of relation samples generated from the same sentence are quite similar, and the only difference is that some of the word position representations between these relation samples are different, which may not be able to show sufficient sample differentiation. In addition, attentive pooling does not extract the most significant features like max-pooling, which may lead to relative deficiencies in distinguishing model similar samples. We will try to analysis and validate these speculations in our future work. 

\subsubsection{F1 score vs. distance}
\begin{figure*}[!hbt]
	\centering
	\includegraphics[width=\textwidth]{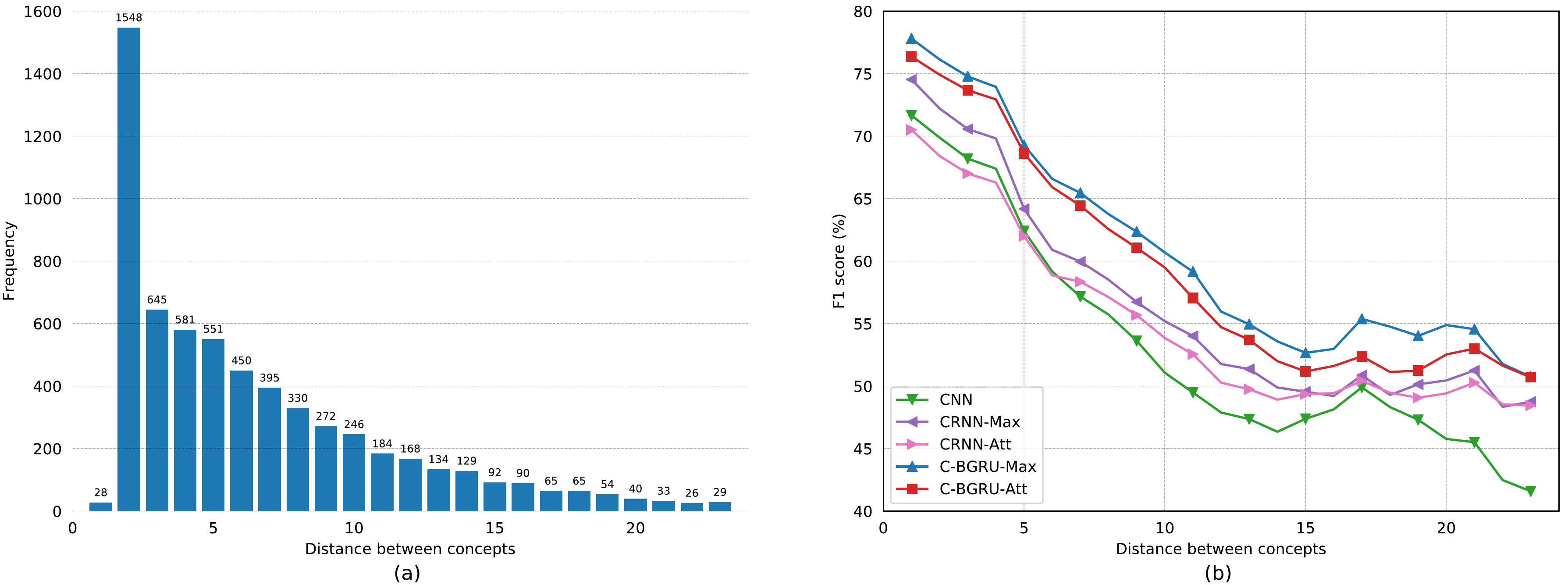}
	\caption{\label{fig:distance} The frequency distribution of the distance between concepts in the 2010 i2b2/VA relation dataset (a) and F1 score comparisons over different distances (b). The ``distance'' means the difference in word position between two concepts in the relation sample.}
\end{figure*}

\begin{figure*}[!hbt]
	\centering
	\includegraphics[width=\textwidth]{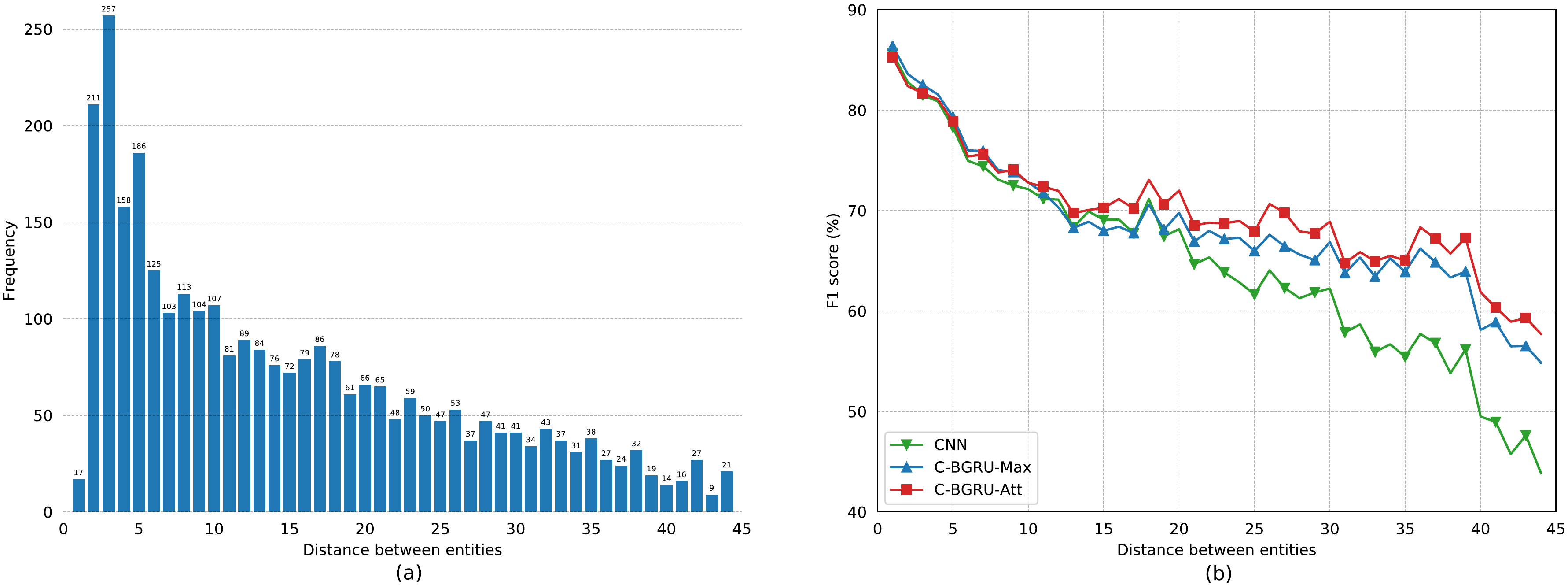}
	\caption{\label{fig:distance_ct} The frequency distribution of the distance between entities in the WI relation dataset (a) and F1 score comparisons over different distances (b). The ``distance" means the difference in word position between two entities in the relation sample.}
\end{figure*}

Figure~\ref{fig:distance}a and \ref{fig:distance_ct}a show the frequency distribution of different distances in the two datasets, and Figure~\ref{fig:distance}b and \ref{fig:distance_ct}b depict the trend of the F1 score as the distance increases. The F1 score is the average value of the relation samples belonging to the distance window $[d - 2, d + 2]$. In order to ensure the reliability of the evaluation, the maximum distance value with a statistic greater than 20 is selected as the truncation of the distance value. On the 2010 i2b2/VA relation dataset, C-BGRU-Max and C-BGRU-Att outperform the baselines over all distances. On the WI relation dataset, C-BGRU-Max and CNN do not show significant differences when the distance is less than 20, but as the distance increases, the performance gap gradually expands. These results verifies that our model has the ability to learn long-term dependencies and this information works in the relation classification task.

\section{Related Work}


Before deep learning research became popular, statistical machine learning methods were the main approaches in the relation classification task. Most of the researchers in the general and clinical domain focused on feature-based and kernel-based methods \cite{Bunescu2005,hendrickx2009semeval,Rink2010,Zhu2013,Kim2015}.

In recent years, researchers have gradually tried the effect of deep learning methods in the relation classification task and achieved satisfactory results. A variety of deep architectures have been proposed to classify the relations, such as recurrent neural network (MV-RNN) \cite{Socher2012}, CNN with softmax classification \cite{Zeng2014}, factor-based compositional embedding model (FCM) \cite{yu2014factor}, and word embedding-based models \cite{hashimoto-EtAl:2015:CoNLL}. Next, there exist many RNN-based and CNN-based variants. Because the max-pooling operation in CNN models will lose significant linguistic features in a sentence, some researchers introduced dependency trees for this work, e.g., bidirectional long short-term memory networks (BLSTM) \cite{Zhang2015a}, dependency-based neural networks (DepNN) \cite{liu-EtAl:2015:ACL-IJCNLP}, shortest dependency path-based CNN \cite{xu-EtAl:2015:EMNLP1}, long short term memory networks along shortest dependency paths (SDP-LSTM) \cite{Xu2015}, deep recurrent neural networks (DRNN) \cite{xu-EtAl:2016:COLING1}, and jointed sequential and tree-structured LSTM-RNN \cite{miwa-bansal:2016:P16-1}. Although the above studies achieved solid results, further research was devoted to eliminating the dependence on the NLP parser because of its limited performance. \citet{DBLP:conf/acl/SantosXZ15} proposed a new pairwise ranking loss function, and only two class representations were updated in every training round. Similarly, \citet{Wang2016} introduced a pairwise margin-based loss function and multi-level attention mechanism and achieved the new state-of-the-art results for relation classification.

More recently, neural network methods have show promising performance for relation classification on clinical records. \citet{Sahu2016Relation} proposed a multiple-filter CNN with some linguistic features, and experiments on the 2010 i2b2/VA relation dataset verified the effectiveness of the neural network model for medical relation classification. \citet{Raj2017} trained a two-layer model by feeding short phrase features extracted by a bidirectional LSTM layer into CNN, and the model performed better than CNN on relation samples where the distance between the medical concepts are large. Different from \citet{Raj2017}'s study, we think n-gram features and sequential correlations among them are the key to relation classification, so we explore another unified architecture that utilizes the strengths of CNN and RNN simultaneously.

\section{Conclusion}

In this paper, we present a unified architecture based on the combination of CNN and RNN to classify medical relations in English and Chinese clinical records. 
Our model captures long-term dependencies of phrase-level features through a bidirectional GRU layer and this information improves model performance.
To the best of our knowledge, this is the first time that neural network methods have been used to classify relations in Chinese clinical text. Experiments show that the proposed model achieves a significant improvement over comparable methods on the 2010 i2b2/VA relation dataset and the WI relation dataset.

\section*{Acknowledgments}
The authors would like to thank all the anonymous reviewers for their insightful comments. We would also like to thank the data support from the 2010 i2b2/VA challenge.

\bibliography{2018he}
\bibliographystyle{AIIM}

\end{document}